\documentclass[article]{jss}

\usepackage{thumbpdf, lmodern, amsmath, listings, graphicx, adjustbox, amssymb}

\lstdefinestyle{mystyle}{
    basicstyle=\ttfamily\footnotesize,
    breakatwhitespace=false,         
    breaklines=true,                 
    captionpos=b,                    
    keepspaces=true,                 
    numbers=left,                    
    numbersep=5pt,                  
    showspaces=false,                
    showstringspaces=false,
    showtabs=false,                  
    tabsize=2
}

\makeatletter
\def\test@relax{\relax}
\let\save@fnum@lstlisting\fnum@lstlisting
\def\fnum@lstlisting{%
    \save@fnum@lstlisting
    \ifx\lst@caption\test@relax\expandafter\@gobble\fi
    }
\makeatother

\author{Trent Henderson\\ The University of Sydney
\And Ben D. Fulcher\\ The University of Sydney}
\Plainauthor{Trent Henderson, Ben Fulcher}

\title{Feature-Based Time-Series Analysis in R using the theft Package}
\Shorttitle{The \pkg{theft} Package for \proglang{R}}

\Abstract{
  Time series are measured and analyzed across the sciences.
  One method of quantifying the structure of time series is by calculating a set of summary statistics or `features', and then representing a time series in terms of its properties as a feature vector.
  The resulting feature space is interpretable and informative, and enables conventional statistical learning approaches, including clustering, regression, and classification, to be applied to time-series datasets.
  Many open-source software packages for computing sets of time-series features exist across multiple programming languages, including \pkg{catch22} (22 features: \proglang{Matlab}, \proglang{R}, \proglang{Python}, \proglang{Julia}), \pkg{feasts} (42 features: \proglang{R}), \pkg{tsfeatures} (63 features: \proglang{R}), \pkg{Kats} (40 features: \proglang{Python}), \pkg{tsfresh} (779 features: \proglang{Python}), and \pkg{TSFEL} (390 features: \proglang{Python}).
  However, there are several issues: (i) a singular access point to these packages is not currently available; (ii) to access all feature sets, users must be fluent in multiple languages; and (iii) these feature-extraction packages lack extensive accompanying methodological pipelines for performing feature-based time-series analysis, such as applications to time-series classification.
  Here we introduce a solution to these issues in the form of a statistical software package for \proglang{R} called \pkg{theft}: Tools for Handling Extraction of Features from Time series.
  \pkg{theft} is a unified and extendable framework for computing features from the six open-source time-series feature sets listed above.
  It also includes a suite of functions for processing and interpreting the performance of extracted features, including extensive data-visualization templates, low-dimensional projections, and time-series classification operations.
  With an increasing volume and complexity of large time-series datasets in the sciences and industry, \pkg{theft} provides a standardized framework for comprehensively quantifying and interpreting informative structure in time series.
}

\Keywords{time-series analysis, time-series features, \proglang{R}, machine learning}
\Plainkeywords{time-series analysis, time-series features, R, machine learning}

\Address{
  Trent Henderson\\
  Dynamics and Neural Systems Group\\
  School of Physics\\
  The University of Sydney\\
  Camperdown NSW 2006, Australia\\
  E-mail: \email{then6675@uni.sydney.edu.au}
}

\begin{document}

\section{Introduction}
\label{sec:intro}

Taking repeated measurements of some quantity through time, forming a time series, is common across the sciences and industry.
The types of time series commonly analyzed are diverse, ranging from signals from an electroencephalogram \citep{westEvaluationComparisonEEG1999}, CO$_2$ concentration in the atmosphere \citep{kodraExploringGrangerCausality2011}, light-curves from distant stars \citep{barbaraClassifyingKeplerLight2022}, and the number of clicks on a webpage \citep{kaoPredictionRemainingTime}.
We can ask many different questions about such data, for example: (i) ``can we distinguish the dynamics of brain disorders from neurotypical brain function?''; (ii) ``can we classify different geospatial regions based on their temporal CO$_2$ concentration''; or (iii) ``can we classify new stars based on their light curves?''.
One approach to answering such questions is to capture properties of each time series and use that information to train a classification algorithm.
This can be achieved by extracting from each time series a set of interpretable summary statistics or `features'.
Using this procedure, a collection of univariate time series can be represented as a time series $\times$ feature matrix which can be used as the basis for a range of conventional statistical learning procedures \citep{fulcherHighlyComparativeTimeseries2013, fulcherFeatureBasedTimeSeriesAnalysis2018}.

The range of time-series analysis methods that can be used to define time-series features is vast, including properties of the distribution, autocorrelation function, stationarity, entropy, methods from the physics nonlinear time-series analysis literature \citep{fulcherHighlyComparativeTimeseries2013}.
Because features are direct outputs of a mathematical operation, and are often tightly linked to underlying theory (e.g., Fourier analysis or information theory), they can yield interpretable understanding of patterns in time series and the processes that produce them---information that can guide further investigation.
The first work to organize these methods from across the interdisciplinary literature encoded thousands of diverse time-series analysis methods as features and compared their behavior on a wide range of time series \citep{fulcherHighlyComparativeTimeseries2013}.
The resulting interdisciplinary library of thousands of time-series features has enabled new ways of doing time-series analysis, including the ability to discover high-performing methods for a given problem in a systematic, data-driven way through large-scale comparison (overcoming the subjective and time-consuming task of selecting methods manually) \citep{fulcherHighlyComparativeFeaturebased2014}.
This approach has been termed `highly comparative time-series analysis', and has been implemented in the \proglang{Matlab} software \pkg{hctsa}, which computes $>7700$ time-series features \citep{fulcherHctsaComputationalFramework2017}.
The approach of automated discovery provided by \pkg{hctsa} has been applied successfully to many scientific problems, such as classifying zebra finch motifs across different social contexts \citep{paulBehavioralDiscriminationTimeseries2021}, classifying cord pH from fetal heart-rate dynamics \citep{fulcherHighlyComparativeFetal2012}, and classifying changes in cortical dynamics from manipulating the firing of excitatory and inhibitory neurons \citep{markicevicCorticalExcitationInhibition2020}. 
While \pkg{hctsa} is comprehensive in its coverage of time-series analysis methods, calculating all of its features on a given dataset is computationally expensive and it requires access to the proprietary \proglang{Matlab} software, limiting its broader use.

The past decade has seen the development of multiple software libraries that implement different sets of time-series features across a range of open-source programming languages.
Here, we focus on the following six libraries:

\begin{itemize}
\item \pkg{catch22} (\proglang{C}, \proglang{Matlab}, \proglang{R}, \proglang{Python}, \proglang{Julia}) computes a representative subset of 22 features from \pkg{hctsa} \citep{lubbaCatch22CAnonicalTimeseries2019}.
The $>$7700 features in \pkg{hctsa} were applied to 93 time-series classification tasks to retain the smallest number of features that maintained high performance on these tasks while also being minimally redundant with each other, yielding the \pkg{catch22} set.
\pkg{catch22} was coded in \proglang{C} for computational efficiency, with wrappers for \proglang{Matlab}, and packages for: \proglang{R}, as \pkg{Rcatch22} \citep{Rcatch22_pkg}; \proglang{Julia}, as \pkg{Catch22.jl} \citep{catch22jl_pkg}; and \proglang{Python}, as \pkg{pycatch22}.
The construction of the \pkg{catch22} feature set focused on dynamical properties, but users can also include the mean and standard deviation of time series in addition to the regular 22 features (yielding the 24-feature set termed `catch24').

\item \pkg{tsfeatures} (\proglang{R}) is the most prominent package for computing time-series features in \proglang{R} \citep{tsfeatures_pkg}.
The 63 features in \pkg{tsfeatures} include techniques commonly used by econometricians and forecasters, such as crossing points, seasonal and trend decomposition using Loess (STL; \cite{clevelandSeasonalTrend1990}), autoregressive conditional heteroscedasticity (ARCH) models, unit-root tests, and sliding windows. 
\pkg{tsfeatures} also includes a small subset of features from \pkg{hctsa} that were previously used to organize tens of thousands of time series in the {\it CompEngine} time-series database \citep{fulcherSelforganizingLivingLibrary2020}.

\item \pkg{feasts} (\proglang{R}) shares a subset of the same features as \pkg{tsfeatures}, computing a total of 42 features.
However, the scope of \pkg{feasts} as a software package is larger: it is a vehicle to incorporate time-series features into the software ecosystem known as the \pkg{tidyverts}\footnote{\url{https://tidyverts.org}}---a collection of packages for time series that follow tidy data principles \citep{wickhamTidyData2014}.
This ensures alignment with the broader and popular \pkg{tidyverse} collection of packages for data wrangling, summarization, and statistical graphics \citep{wickhamWelcomeTidyverse2019}.
\pkg{feasts} also includes functions for producing graphics, but these are largely focused on exploring quantities of interest in econometrics, such as autocorrelation, seasonality, and STL decomposition.

\item \pkg{tsfresh} (\proglang{Python}) includes 779 features that measure properties of the autocorrelation function, entropy, quantiles, fast Fourier transforms, and distributional characteristics \citep{christDistributedParallelTime2017}.
\pkg{tsfresh} also includes a built-in feature filtering procedure, FeatuRe Extraction based on Scalable Hypothesis tests (FRESH), that uses a hypothesis-testing process to control the percentage of irrelevant extracted features \citep{christTimeSeriesFeatuRe2018}.
\pkg{tsfresh} has been used widely to solve time-series problems, such as anomaly detection in Internet-of-Things streaming data \citep{yangAnomalyDetectionAlgorithm2021} and sensor-fault classification \citep{liuSensorFaultsClassification2020}.

\item \pkg{TSFEL} (\proglang{Python}) contains 390 features that measure properties associated with distributional characteristics, the autocorrelation function, fast Fourier transforms, spectral quantities, and wavelets \citep{barandasTSFELTimeSeries2020}. 
\pkg{TSFEL} was initially designed to support feature extraction of inertial data---such as data produced by human wearables---for the purpose of activity detection and rehabilitation.

\item \pkg{Kats} (\proglang{Python}), developed by Facebook Research, contains a broad range of time-series functionality, including operations for forecasting, outlier and property detection, and feature calculation \citep{Kats}.
The feature-calculation module of \pkg{Kats} is called \code{TSFeatures} and includes 40 features (30 of which are based on \proglang{R}'s \pkg{tsfeatures} package).
\pkg{Kats} includes features associated with crossing points, STL decomposition, sliding windows, autocorrelation and partial autocorrelation, and Holt--Winters methods for detecting linear trends.
\end{itemize}

The six sets vary over several orders of magnitude in their computation time, and exhibit large differences in both within-set feature redundancy---how correlated features are within a given set---and between-set feature redundancy---how correlated, on average, features are between different pairwise comparisons of sets \citep{hendersonEmpiricalEvaluationTimeSeries2021}.
While each set contains a range of features that could be used to tackle time-series analysis problems, there are currently no guidelines for selecting an appropriate feature set for a given problem, nor methods for combining the different strengths of all sets.
Performance on a given time-series analysis task depends on the choice of the features that are used to represent the time series, highlighting the importance of being able to easily compute many different features from across different feature sets.
Furthermore, following feature extraction, there is no set of visualization and analysis templates for common feature-based problem classes, such as feature-based time-series classification (like the tools provided in \pkg{hctsa} \cite{fulcherHctsaComputationalFramework2017}).
Here we present a solution for these challenges in the form of an open-source package for \proglang{R} called \pkg{theft}: Tools for Handling Extraction of Features from Time series. 

\section[The theft package for R]{The \pkg{theft} package for \proglang{R}}
\label{sec:theft}

\pkg{theft} unifies the six free and open-source feature sets described in Section \ref{sec:intro}, thus overcoming barriers in using diverse feature sets developed in different software environments and using different syntax.
\pkg{theft} also provides an extensive analytical pipeline and statistical data visualization templates similar to those found in \pkg{hctsa} for understanding feature behavior and performance.
Such pipelines and templates do not currently exist in the free and open-source setting, making \pkg{theft} an invaluable tool for both computing and understanding features.
While there is some software support for computing features in a consistent setting (such as in \pkg{tsflex} \cite{vanderdoncktTsflexFlexibleTime2022}, which also provides sliding window extraction capability), such software is limited to specifying the functional form of individual time-series features rather than automatically accessing every feature contained in different sets.

\begin{figure}[t!]
  \centering
  \includegraphics[width = 0.5\textwidth]{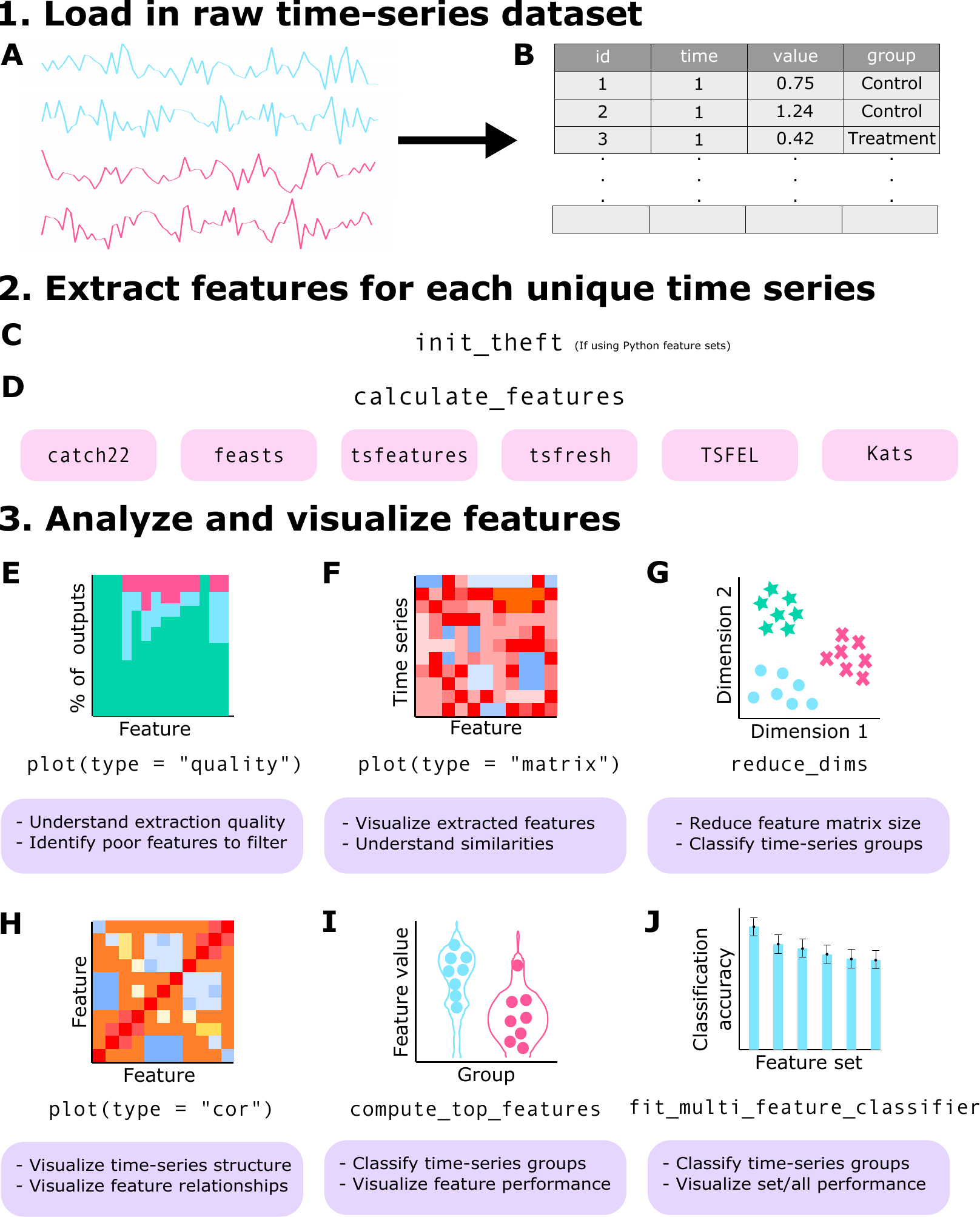}
  \caption{
  \textbf{\pkg{theft} implements a workflow for extracting features from univariate time series and processing and analyzing the results.}
  First, a time-series dataset (\textbf{A}) is converted into a tidy (`long') data frame (\textbf{B}) with variable names for unique identifiers, time-point indices, values, and group labels (e.g., in the case of classification problems).
  If one of the feature sets selected is a \proglang{Python} library, the user can point \proglang{R} to the \proglang{Python} version containing the installed software (\textbf{C}). 
  One or more feature sets are then computed on the dataset (\textbf{D}).
  A range of statistical analysis and data visualization functionality is also implemented, including:
  (\textbf{E}) feature quality assessment (e.g., understanding the proportion of non-\code{NA} values by feature);
  (\textbf{F}) normalized time series $\times$ feature matrix visualization;
  (\textbf{G}) low-dimensional projections of the feature space; and
  (\textbf{H}) normalized feature $\times$ feature correlation matrix visualization.
  Functionality is also provided for time-series classification (a common application of feature-based time-series analysis), including:
  (\textbf{I}) understanding the most discriminative individual features; and
  (\textbf{J}) fitting and evaluating classifiers with more than one feature as input.
  \label{fig:workflow}
  }
\end{figure}

The functionality provided by \pkg{theft} is summarized in Fig.~\ref{fig:workflow} and broadly follows the feature-based time-series analysis workflow of \pkg{hctsa} \citep{fulcherHctsaComputationalFramework2017}.
The workflow begins with a time-series dataset (Fig.~\ref{fig:workflow}A) that is converted to a tidy format (Fig.~\ref{fig:workflow}B).
If any of the \proglang{Python} feature sets are to be used, the \proglang{Python} environment containing the installed software is passed to \pkg{theft} using \code{init\_theft} (Fig.~\ref{fig:workflow}C).
Time-series features are then extracted (Fig.~\ref{fig:workflow}D).
The user can pass the extracted features into a range of statistical and visualization functions to derive interpretable understanding of the informative patterns in their dataset (Figs~\ref{fig:workflow}E--J). 
Importantly, \pkg{theft} uses R's \code{S3} object-oriented programming system, meaning classes and their methods are defined to ensure usability. 
Classes are defined for feature calculation objects (\textbf{D}; \code{feature\_calculations}) and low dimensional projection objects (\textbf{G}; \code{low\_dimension}). 
Almost every \pkg{theft} function takes at least an object of one of these classes as input. 
This means that generic methods, such as \code{plot} can be called on the object, where additional arguments to the generic function such as \code{type} declares the type of graphic that is produced. 
These methods are discussed in detail in the following sections.

In this paper, we demonstrate how \pkg{theft} can be used to tackle a time-series classification problem, using the Bonn University electroencephalogram (EEG) dataset as a case study \citep{andrzejakIndicationsNonlinearDeterministic2001}.
The dataset contains 500 time series, with 100 time series each from five labeled classes: (i) awake with eyes open (labeled `eyesOpen'); (ii) awake with eyes closed (`eyesClosed'); (iii) epileptogenic zone (`epileptogenic'); (iv) hippocampal formation of the opposite hemisphere of the brain (`hippocampus'); and (v) seizure activity (`seizure').
Note that classes (i) and (ii) are from healthy volunteers, while classes (iii), (iv), and (v) are from a presurgical diagnosis archive.
This dataset was chosen as a demonstrative example because it has been widely studied as a time-series classification problem, and prior studies have focused on properties of the dynamics that accurately distinguish the classes---which is well-suited to the feature-based approach. 
For example, an analysis using \pkg{hctsa} revealed that seizure recordings are characterized most notably by higher variance, as well as lower entropy, lower long-range scaling exponents, and many other differences \citep{fulcherHighlyComparativeTimeseries2013}. 
Further, it was also found that 172 individual features within \pkg{hctsa} could distinguish between healthy EEGs and seizures using a 10-fold cross-validation linear classifier with $>95\%$ accuracy, with eight of these features achieving $>98.75\%$ --- exceeding previous results which used operations from the discrete wavelet transform as inputs to a support vector machine classifier \citep{subasiEEGSignalClassification2010}.

\subsection{Extracting features}
\label{sec:calcs}

In feature-based time-series analysis, each univariate time series in a dataset is represented as a feature vector, such that the dataset can be represented as a time series $\times$ feature data matrix.
Any single feature set, or combination of multiple feature sets, can be computed for a given time-series dataset with the \pkg{theft} function \code{calculate\_features}.
An example call which extracts features from all six sets is shown in Listing~\ref{lst:calc_feats}.
Note that throughout this paper we will present the code to execute each piece of analysis, but will largely omit the extensive amount of optional arguments for clarity.
The vignette and function documentation that comes with every download of \pkg{theft} contains detailed information regarding these arguments \citep{theft_pkg}.
The call in Listing~\ref{lst:calc_feats} takes a tidy data frame that contains the time-series data (\code{tmp} in this example), takes the relevant user-specified columns, and computes the time series $\times$ feature matrix for the specified feature set(s).
The output of this function is an S3 object of class \code{feature\_calculations}, which in this example is stored in the R environment as \code{feature\_matrix}.
Within this object is a data frame which contains five columns if the dataset is labeled (as in time-series classification), and four otherwise: \code{id} (unique identifier for each time series), \code{names} (feature name), \code{values} (feature value), \code{method} (feature set), and \code{group} (class label, if applicable).
This output structure ensures that, regardless of the feature set selected, the resulting object is always of the same format and can be used with the rest of \pkg{theft}'s functions without manual data reshaping.
For the Bonn EEG dataset, calculating features for all sets in \pkg{theft} takes several hours.

\begin{lstlisting}[language=R, label={lst:calc_feats}, caption=\relax]
  all_features <- calculate_features(
                        data = tmp, 
                        id_var = "id", 
                        time_var = "timepoint", 
                        values_var = "values", 
                        group_var = "group",
                        feature_set = c("catch22", "feasts", "tsfeatures", 
                                        "tsfresh", "TSFEL", "Kats")
\end{lstlisting}

\subsection{Assessing feature extraction quality}
\label{sec:quality}

Not all features return real-valued outputs for all time series, meaning non-numeric values, such as \code{NaN} or \code{Inf/-Inf}, or even missing outputs or errors, can be returned as a result of feature extraction.
For example, a feature which computes the variance across multiple 200-sample time-series windows cannot be computed for time series shorter than 200 samples (as there are not enough samples to form even a single window).
For effective quality control, it is important to visualize numeric and non-numeric outputs following feature extraction, which is implemented in \pkg{theft} by calling \code{plot} on a \code{feature\_calculations} object and specifying a plot type of \code{"quality"}, as shown in Listing~\ref{lst:quality_plot}.
This graphic plots the proportion of values in each that are numeric, \code{NaN/NA}, or \code{Inf/-Inf} for each feature as bars. 

\begin{lstlisting}[language=R, label={lst:quality_plot}, caption=\relax]
  plot(all_features, type = "quality")
\end{lstlisting}

\subsection{Normalizing features}
\label{sec:norms}

Different features vary over very different ranges; e.g., features that estimate $p$-values from a hypothesis test vary over the unit interval, whereas a feature that computes the length of a time series can take (often large) positive integers.
These differences in scale can complicate the visualization of feature behavior and the construction of statistical learning algorithms involving diverse features.
To overcome these limitations, a common pre-processing step involves scaling all features.
\pkg{theft} includes four such methods for converting a set of raw feature values, $\mathbf{x}$, to a normalized version, $\mathbf{z}$:

\begin{enumerate}
  \item $z$-score: $z_i = \frac{x_i - \mu}{\sigma}$,
  \item linear scaling to unit interval: $z_i = \frac{x_i - \min(\mathbf{x})}{\max(\mathbf{x}) - \min(\mathbf{x})}$,
  \item sigmoid: $z_i = \left[1 + \exp(-\frac{x_i - \mu}{\sigma})\right]^{-1}$,
  \item and outlier-robust sigmoid: $z_i = \left[1 + \exp\left(-\frac{x_i - \mathrm{median}(\mathbf{x})}{\mathrm{IQR}(\mathbf{x})/{1.35}}\right)\right]^{-1}$,
\end{enumerate}
where $\mu$ is the mean, $\sigma$ is the standard deviation, and $\mathrm{IQR}(\mathbf{x})$ is the interquartile range of $\mathbf{x}$.
All four transformations end with a linear rescaling to the unit interval.
The outlier-robust sigmoid transformation, introduced in \citet{fulcherHighlyComparativeTimeseries2013}, can be helpful in normalizing feature-value distributions with large outliers.
Normalization is an option in each of the core analysis and visualization functions within \pkg{theft}, but users can also perform normalization outside of these functions on individual feature vectors or \code{feature\_calculations} objects (i.e., the entire feature matrix) using the \code{normalise} function which automatically detects the input object class.
An example for \code{feature\_calculations} objects is shown in Listing~\ref{lst:norms}.

\begin{lstlisting}[language=R, label={lst:norms}, caption=\relax]
  normed <- normalise(all_features, method = "RobustSigmoid")
\end{lstlisting}

\subsection{Visualizing the feature matrix}
\label{sec:datamatrix}

\begin{figure}[t!]
  \centering
  \includegraphics{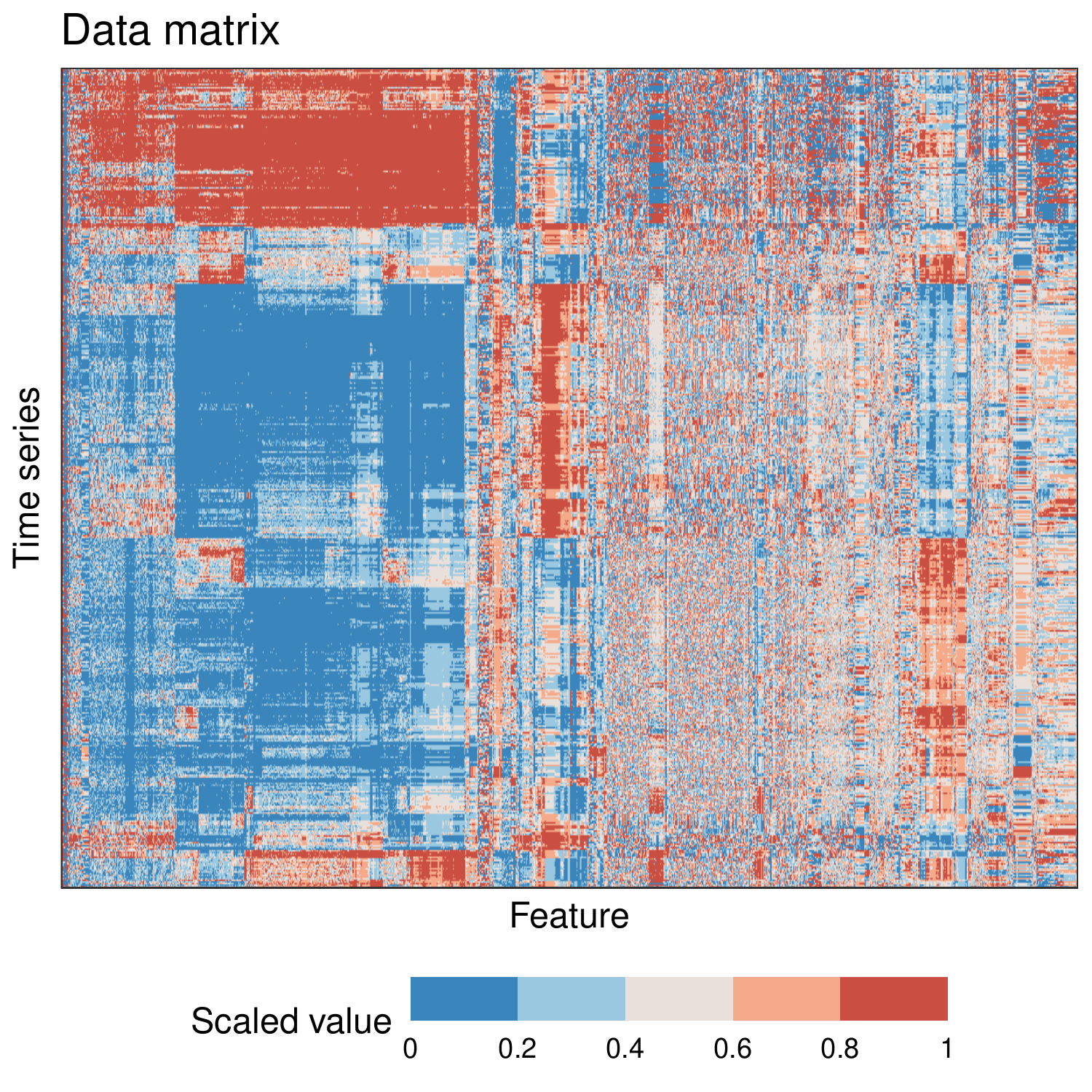}
  \caption{\label{fig:all_features}
  \textbf{A time series $\times$ feature matrix heatmap produced by generating a matrix plot on the \code{feature\_calculations} object.}
  Extracted feature vectors for each time series (500) in the Bonn EEG dataset using all six feature sets in \pkg{theft} (1253 features in total, after filtering out 63 features with \code{NaN} values) are represented as a heatmap.
  Similar features (columns) and time series (rows) are positioned close to each other using (average) hierarchical clustering.
  Each tile is a normalized value for a given time series and feature.
  This plot is generated by the code in Listing~\ref{lst:feat_mat_plot}.
  }
\end{figure}

A hallmark of large-scale feature extraction is the ability to visualize the intricate patterns of how different scientific algorithms behave across a time-series dataset.
This can be achieved in \pkg{theft} using by specifying \code{type = "matrix"} when calling \code{plot} on a \code{feature\_calculations} object to produce a heatmap of the time series (rows) $\times$ feature matrix (columns) which organizes the rows and columns to help reveal interesting patterns in the data.
The plot of the combination of all six open feature sets for the Bonn EEG dataset is shown in Fig.~\ref{fig:all_features}, with the code displayed in Listing~\ref{lst:feat_mat_plot}.
We can see some informative structure in this graphic, including many groups of features with similar behavior on this dataset (i.e., columns with similar patterns), indicating substantial redundancy across the joint set of features \citep{hendersonEmpiricalEvaluationTimeSeries2021}.
The top block of 100 rows, which visually have the most distinctive properties, were found to correspond to time series from the ``seizure'' class, indicating the ability of this large combination of time-series features to meaningfully structure the dataset.

In matrix plots in \pkg{theft}, hierarchical clustering is used to reorder rows and columns so that time series (rows) with similar properties are placed close to each other and features (columns) with similar behavior across the dataset are placed close to each other---where similarity in behavior is quantified using Euclidean distance in both cases \citep{dayEfficientAlgorithmsAgglomerative1984}.

Default settings within \code{plot} enable users to easily generate outputs in a single line of code, but more advanced users may seek to tweak the optional arguments.
For example, different linkage algorithms for hierarchical clustering can be controlled supplied to \code{clust\_method}, which uses average linkage as a default, and different rescaling methods can be supplied to the \code{method} argument, which defaults to \code{"z-score"}.

\begin{lstlisting}[language=R, label={lst:feat_mat_plot}, caption=\relax]
  plot(all_features, type = "matrix")
\end{lstlisting}

\subsection{Projecting low-dimensional feature-spaces}
\label{sec:lowdim}

Low-dimensional projections are a useful tool for visualizing the structure of high-dimensional datasets in low-dimensional spaces.
Here we are interested in representing a time-series dataset in a two-dimensional projection of the high-dimensional feature space, which can reveal structure in the dataset, including how different labeled classes are organized.
For linear dimensionality reduction techniques---such as principal components analysis (PCA) \citep{PrincipalComponentAnalysis2002}---the results can be visualised in two dimensions as a scatterplot, where the principal component (PC) that explains the most variance in the data is positioned on the horizontal axis and the second PC on the vertical axis, and each time series is represented as a point (colored by its group label in the case of a labeled dataset).
When the structure of a dataset in the low-dimensional feature space matches known aspects of the dataset (such as class labels), it suggests that the combination of diverse time-series features can capture relevant dynamical properties that differ between the classes.
It can also reveal new types of structure in the dataset, like clear sub-clusters within a labeled class, that can guide new understanding of the dataset.
Low-dimensional projections of time-series features have been shown to meaningfully structure time-series datasets---revealing sex and day/night differences in \textit{Drosophila} \citep{fulcherHctsaComputationalFramework2017}, distinguishing types of stars based on their light curves \citep{barbaraClassifyingKeplerLight2022}, and categorizing sleep epochs \citep{decatTraditionalSleepScoring2022}.

In \pkg{theft}, both a linear dimensionality reduction method---PCA---and a nonlinear dimensionality reduction method---$t$-distributed stochastic neighbour embedding ($t$-SNE) \citep{maatenVisualizingDataUsing2008}---are included.
While many dimensionality-reduction algorithms exist \citep{sorzanoSurveyDimensionalityReduction2014}, here we selected just these two to minimize package dependencies.
Time-series datasets can be projected in low-dimensional feature spaces in \pkg{theft} using the \code{reduce\_dims} function demonstrated in Listing~\ref{lst:low_dim}.
For $t$-SNE, users can control the \textit{perplexity} hyperparameter using the \code{perplexity} argument. 
The \code{reduce\_dims} function returns an S3 object of class \code{low_dimension}, which can then be passed to \code{plot} which will automatically detect the class. 
The plot method for this object contains only one other argument (\code{show\_covariance}) which is a Boolean specifying whether to draw covariance ellipses for each group in the scatterplot (if a grouping variable is detected).

\begin{lstlisting}[language=R, label={lst:low_dim}, caption=\relax]
  low_dim_calc <- reduce_dims(all_features, method = "MinMax", low_dim_method = "t-SNE", perplexity = 15)
\end{lstlisting}

\begin{figure}[t!]
  \centering
  \includegraphics{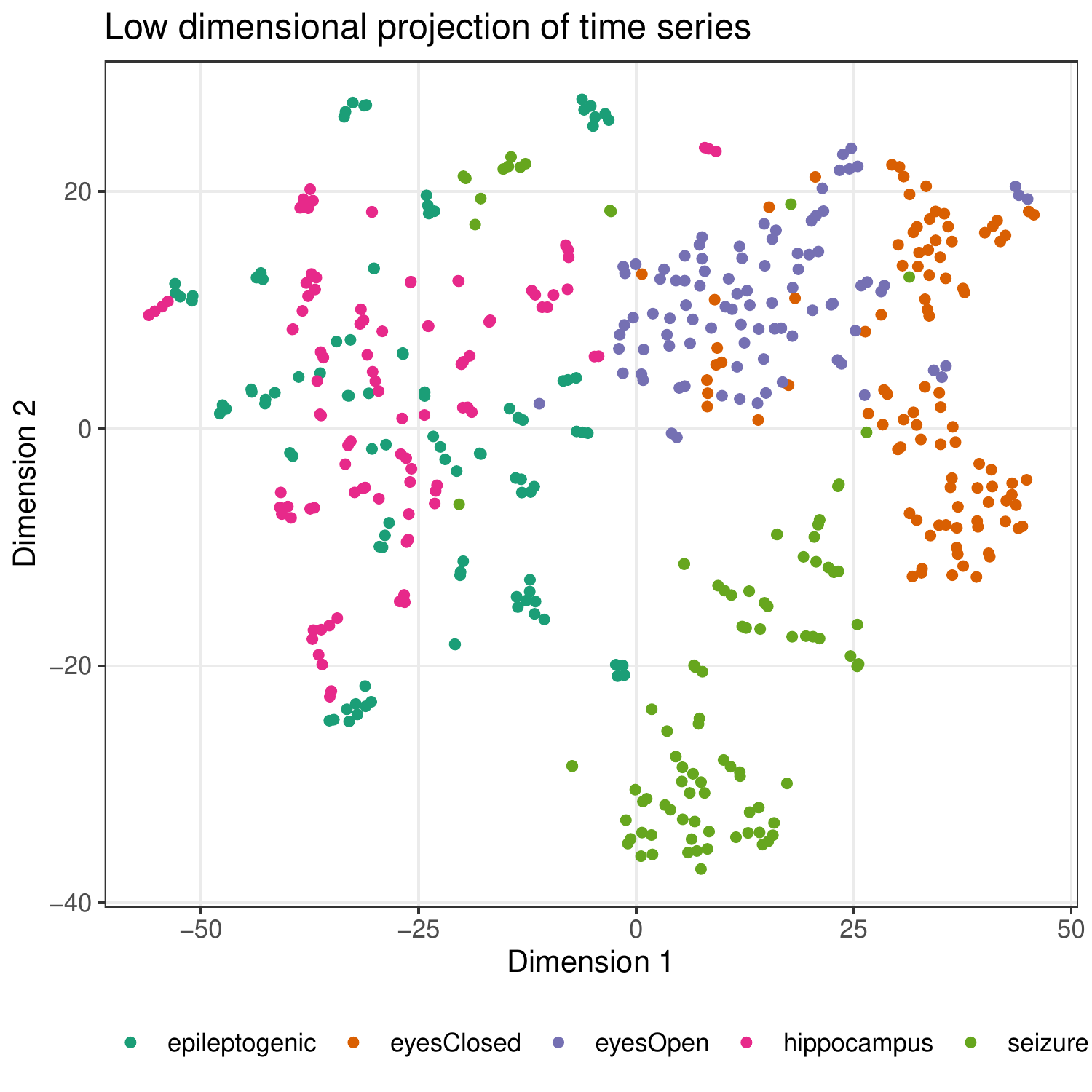}
  \caption{\label{fig:tsne}
  \textbf{Low-dimensional projection of the Bonn EEG dataset using \pkg{theft}}.
  Using $t$-SNE with perplexity 15, the high-dimensional feature space of over $1200$ features is projected into two dimensions.
  Each point represents a time series which is colored according to its class label.
  Time series that are located close in this space have similar properties, as measured by the six feature sets in \pkg{theft}.
  This plot is generated by calling \code{plot(low\_dim\_calc, show\_covariance = FALSE)} after the code in Listing~\ref{lst:low_dim}.
  }
\end{figure}

The low-dimensional projection plot for the Bonn EEG dataset (using $t$-SNE and all non-\code{NaN} features across the six feature sets included in \pkg{theft}) is shown in Fig.~\ref{fig:tsne} with perplexity 10, as produced by the code in Listing~\ref{lst:low_dim}.
The low-dimensional projection (formed from $>1200$ features in \pkg{theft}) meaningfully structures the labeled classes of the dataset.
Specifically, two of the presurgical diagnosis classes---``epileptogenic'' (epileptogenic zone) and ``hippocampus'' (hippocampal formation of the opposite hemisphere of the brain)---appear to exhibit considerable overlap in the projected space, while the two healthy volunteer classes ``eyesOpen'' (awake state with eyes open) and ``eyesClosed'' (awake state with eyes closed) occupy space further away from the other classes but closer to each other.
The ``seizure'' class occupies a space largely separate from the other four classes in the projection, consistent with its distinctive dynamics \citep{fulcherHighlyComparativeTimeseries2013}.

\subsection{Constructing classifiers with multiple features}
\label{sec:multi_classification}

Combinations of complementary, discriminative features can often be used to construct accurate time-series classifiers \citep{fulcherHighlyComparativeFeaturebased2014}.
Drawing on computed time-series features (that may derive from one or more existing feature sets), \pkg{theft} can fit and evaluate classifiers using the \code{fit\_multi\_feature\_classifier} function.
This allows users to evaluate the relative performance of each feature set, of the combination of all sets, or any other combination of features.
Providing easy access to a range of classification algorithms and accompanying inferential tools (such as null permutation testing to obtain $p$-values) through \code{fit\_multi\_feature\_classifier} allows users to compare sets of features to better understand the most accurate feature sets for a given time-series classification problem.
The code presented in Listing~\ref{lst:mf_classifier} provides an example usage for the Bonn EEG dataset with a linear support vector machine (SVM) classifier.

\begin{lstlisting}[language=R, label={lst:mf_classifier}, caption=\relax]
  mf_results <- fit_multi_feature_classifier(
                    data = all_features, 
                    by_set = TRUE, 
                    test_method = "svmLinear",
                    use_k_fold = TRUE, 
                    num_folds = 10)
\end{lstlisting}

The \code{fit\_multi\_feature\_classifier} function returns a list object.
If \code{by\_set} is \code{TRUE}, an plot object called \code{FeatureSetResultsPlot} is created and returned in the list which contains a bar plot of classification accuracy for each feature set (if \code{by\_set} is set to \code{FALSE}, all features will be used as predictors in the chosen classification model, ignoring the set they originate from and not returning a plot).

The modeling components of \code{fit\_multi\_feature\_classifier} are executed through a wrapper for the machine-learning package \pkg{caret} \citep{caret}.
This means the extensive list of classification models available in \pkg{caret} can be accessed from \pkg{theft} by specifying the method name in the \code{test\_method} argument.
Prior to fitting a model, \code{fit\_multi\_feature\_classifier} performs two operations:
(i) filtering out features that are constants or contain \code{NaN/NA} or \code{Inf/-Inf} values;
and (ii) re-coding of class labels into syntactically valid names for model objects in \proglang{R}.
The resulting data is then passed into a \pkg{caret} \code{train} operation, where if \code{use\_k\_fold} is set to \code{TRUE}, $k$-fold cross-validation is performed, with the number of folds set by the \code{num\_folds} argument.
All operations produced by \code{fit\_multi\_feature\_classifier} use centering and scaling preprocessing provided by \pkg{caret} and exclude time-series features with near-zero variance after executing this procedure.
The performance metric is specified by \code{use\_balanced\_accuracy} where if \code{FALSE}, mean classification accuracy is used and if \code{TRUE}, balanced mean classification accuracy is used.
Balanced classification accuracy is a useful metric for problems where class imbalances can artificially inflate the accuracy metric (i.e., a classifier might assign the majority or all of the predicted class labels to the class with the most representation in the data).

\begin{figure}[h!]
  \centering
  \includegraphics[width = 0.8\textwidth]{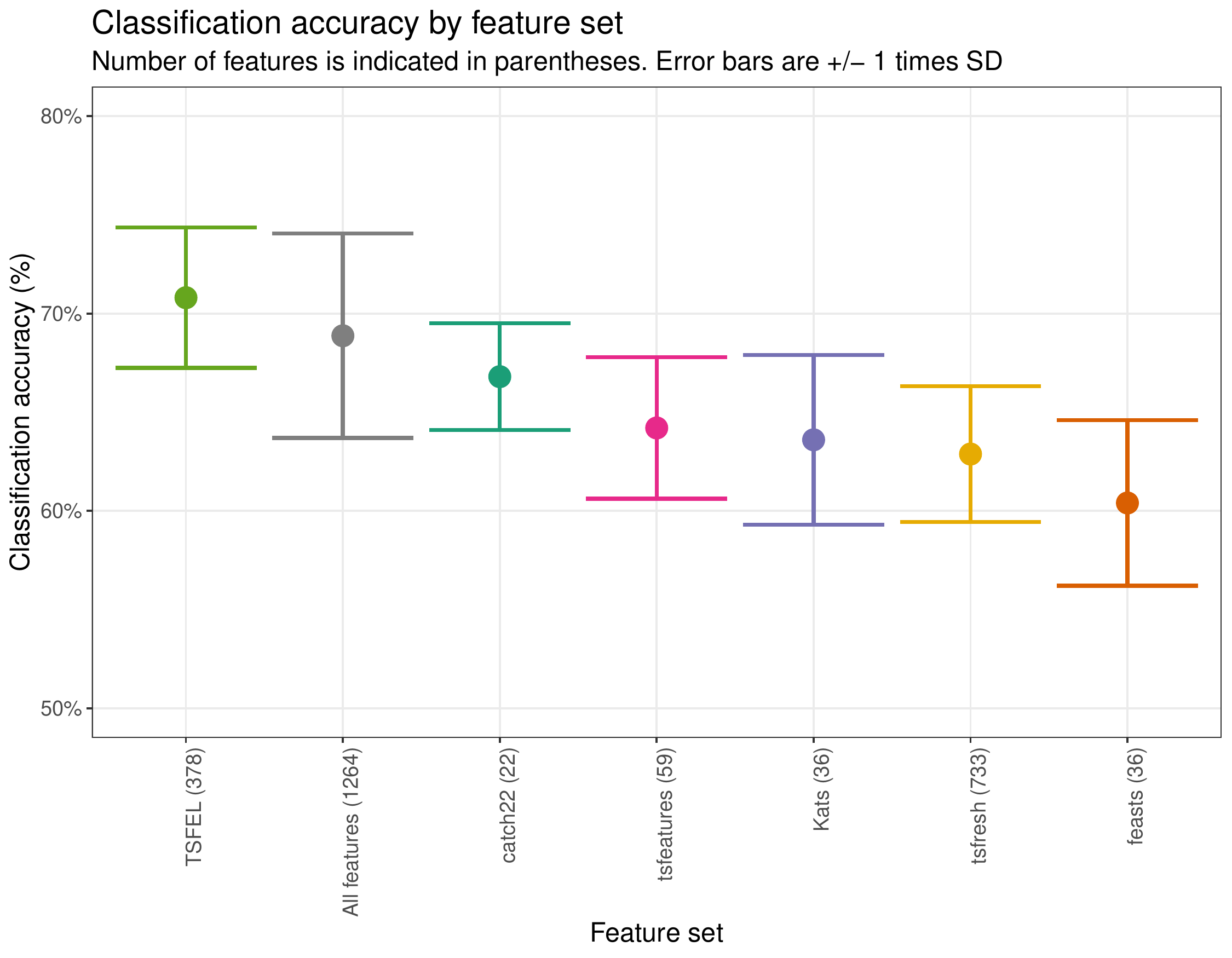}
  \caption{\label{fig:feature-set-comparisons}
  \textbf{Comparison of mean classification accuracy between feature sets in \pkg{theft} for the five-class Bonn EEG classification task}.
  Classification accuracy using a linear SVM is presented for each of the six feature sets in \pkg{theft} as well as the combination of all their features (`All features').
  The number of features retained for analysis after filtering is displayed in parentheses after the feature set name on the horizontal axis which has been sorted from highest to lowest mean accuracy. 
  Mean classification accuracy across the same 10 cross-validation folds is displayed as colored points for each set with $\pm 1$\textit{SD} error bars.
  This plot is generated by the code in Listing~\ref{lst:mf_classifier}.
  }
\end{figure}

The summary \code{FeatureSetResultsPlot} graphic produced by Listing~\ref{lst:mf_classifier} for the Bonn EEG dataset (in which all time series have been $z$-scored, to focus on differences in dynamical properties between the classes) is shown in Fig.~\ref{fig:feature-set-comparisons}.
On this problem, we find that \pkg{TSFEL} (with 378 features after filtering) has the highest mean classification accuracy ($70.8\%$, \textit{SD} = $3.6\%$ over 10 folds) and \pkg{feasts} has the lowest mean accuracy ($60.4\%$, \textit{SD} = $4.2\%$).
The combination of all $>1260$ features from across every set demonstrates accuracy consistent with the top performer (\pkg{TSFEL}), with slightly lower mean performance (potentially due to over-fitting in a higher-dimensional feature space).

\subsubsection{Permutation testing}
\label{sec:perm_testing}

In applications involving small datasets, or when small effects are expected, it is useful to quantify how different the calculated classification performance is from a null setting in which data are classified randomly.
One method for inferring test statistics is to use permutation testing---a procedure that samples a null process many times to form a distribution against which a value of importance (i.e., the classification accuracy result from a model) can be compared to estimate a $p$-value \citep{ojalaPermutationTestsStudying2009}.
In \pkg{theft}, permutation testing is implemented for evaluating classification performance in \code{fit\_multi\_feature\_classifier} through the \code{use\_empirical\_null} argument.
When set to \code{TRUE}, an object is returned in the list called \code{TestStatistics} which contains a data frame of classification accuracy results and associated $p$-values.
Regardless of the options specified to \code{by\_set} and \code{use\_empirical\_null}, an object called \code{RawClassificationResults} is always returned in the overall list object created by \code{fit\_multi\_feature\_classifier}, which contains a data frame of classification accuracy outputs from each model that is trained and evaluated. 

A more detailed version of Listing~\ref{lst:mf_classifier} with the optional arguments specified to execute the null testing procedure is displayed in Listing~\ref{lst:mf_classifier_opts}.

\begin{lstlisting}[language=R, label={lst:mf_classifier_opts}, caption=\relax]
  mf_results <- fit_multi_feature_classifier(
                    data = all_features, 
                    by_set = TRUE, 
                    test_method = "svmLinear",
                    use_empirical_null = TRUE, 
                    null_testing_method = "ModelFreeShuffles",
                    p_value_method = "gaussian", 
                    num_permutations = 10000)
\end{lstlisting}

Two methods for permutation testing are implemented: (i) model-free random shuffles \\(\code{"ModelFreeShuffles"})---from the vector of class labels in the dataset, randomly permutes \code{num\_permutations} random shuffles (permutations) and calculates either mean classification accuracy or balanced accuracy for each shuffled vector against the original vector of class labels; and (ii) null model fits (\code{"NullModelFits"})---fits \code{num\_permutations} models with the same classifier, same input data, and same $k$-fold cross-validation procedure as the main classification model, but trains the null models on randomly shuffled class labels as the target variable.

The model-free procedure (which assumes a null model that produces randomized outputs labels) is very fast, and provides a good approximation for the null distribution obtained using randomized input labels when cross-validation is used (i.e., when all evaluation is performed on unseen data).
From the resulting null distribution, a $p$-value can be assigned to the main model's performance statistic via one of two methods: (i) \code{"gaussian"} and (ii) \code{"empirical"}.
The former takes the mean and standard deviation of the null classification accuracy values, and evaluates the probability of the classification accuracy result of the model fit on the correct class labels against a Gaussian distribution parameterized by this null mean and standard deviation.
The latter simply calculates the proportion of null classification accuracy values that are equal to or greater than the classification accuracy result of the model fit on the correct class labels.
If the null distribution is believed to be approximately Gaussian, then an estimate of potentially small $p$-values can be obtained using the \code{"gaussian"} setting with less samples rather than expending a large computing time to resolve the potential for small $p$-values through permutation testing and frequencies using the \code{"empirical"} setting.

For the full Bonn EEG dataset, with 100 time series per class and strong differences between signals, classification accuracies are far higher than chance level (20\%), and we obtain extremely small $p$-values, confirming the low probability of obtaining such high accuracies by chance.
To more clearly demonstrate the permutation testing functionality, we analyzed a smaller random subsample of the dataset: 14 $z$-scored time series each from the `hippocampus' and `epileptogenic' classes, and fit binary classification models using the code presented in Listing~\ref{lst:mf_classifier_opts}.
The mean classification accuracy of \pkg{feasts} ($64\%$, $p = 0.07$), \pkg{catch22} ($68\%$, $p = 0.07$), and \pkg{TSFEL} ($69\%$, $p = 0.07$) were not statistically significantly different (at $p < 0.05$ with all $p$-values adjusted using the Holm-Bonferroni method for multiple comparisons) from the null distribution, despite their mean accuracy values being above the chance probability of $50\%$.
The remaining sets were statistically significant: \pkg{Kats} ($74\%$, $p = 0.02$), \pkg{tsfresh} ($79\%$, $p = 0.007$), and \pkg{tsfeatures} ($84\%$, $p = 0.001$).

\subsection{Finding and understanding informative individual features}
\label{sec:single_classification}

Fitting models which use multiple features as inputs is often useful for predicting class labels.
However, users are also typically interested in understanding patterns in their dataset, such as interpreting the types of time-series analysis methods that best separate different classes, and the relationships between these top-performing features.
This can be achieved using mass univariate statistical testing of individual features, quantifying their importance either with conventional statistical tests (e.g., $t$-tests, Wilcoxon Rank Sum Tests, and Signed Rank Tests), or with one-dimensional classification algorithms (e.g., linear SVM, random forest classifiers).
\pkg{theft} implements the ability to identify top-performing features in the \code{compute\_top\_features} function, with an example usage for the Bonn EEG dataset (using features from all six packages) shown in Listing~\ref{lst:sf_classifier}.

\begin{lstlisting}[language=R, label={lst:sf_classifier}, caption=\relax]
  top_feature_results <- compute_top_features(
                            data = all_features, 
                            num_features = 40, 
                            test_method = "svmLinear",
                            use_k_fold = TRUE,
                            num_folds = 10,
                            use_empirical_null =  TRUE)
\end{lstlisting}

For two-class problems, users can access the traditional statistical tests by specifying either \code{test\_method = "t-test"}, \code{test\_method = "wilcox"}, or \code{test\_method = \\"BinomialLogistic"} to fit the desired statistical test instead of a \pkg{caret} classification model.
\code{compute\_top\_features} allows users to fit the same set of \pkg{caret} classification models available in \code{fit\_multi\_feature\_classifier} in the one-dimensional space (i.e., the input to the algorithm is values on a single time-series feature), which can be used for two-class or multi-class problems (where traditional two-sample statistical tests cannot be used).

Regardless of the \code{test\_method} used, \code{compute\_top\_features} always returns a list object with three elements:
(i) \code{ResultsTable} (a data frame of either statistical test results and model information or classification results and resulting null test statistics based on whether a \pkg{caret} model was used or not);
(ii) \code{FeatureFeatureCorrelationPlot} (pairwise feature $\times$ feature correlation plot with the correlation method in the \code{cor_method} argument); and
(iii) \code{ViolinPlots} (violin plots for each feature that show each time series as a point, colored and arranged in columns by class, with class-level probability density lines around them).

The \code{ResultsTable} summary of classification accuracy values and $p$-values for the top 40 features on the Bonn EEG classification task (determined by $p$-value of mean classification accuracy against the empirical null) is displayed in Table~\ref{tab:top_feature_table}.
There is representation from all six feature sets, with the majority coming from the two largest sets: \pkg{tsfresh} and \pkg{TSFEL}.
Inspecting the feature names within the table can provide insight into the types of features relevant for the classification problem. 
These features inlcude properties associated with wavelets (e.g., \code{0\_wavelet\_energy\_7}, which measures the area under the squared magnitude of the continuous wavelet transform at scale 7),
variance (e.g., \code{0\_standard\_deviation}, which is the standard deviation of the signal), and autocorrelation (e.g., \code{values\_autocorrelation\_lag\_7}, which measures the value of the autocorrelation function at lag 7). 
In addition, it can also be seen that two of strongest performing features ($53.8\%$ individual classification accuracy) are in fact the same feature implemented in two different sets (\code{SC\_FluctAnal\_2\_rsrangefit\_50\_1\_logi\_prop\_r1} from \pkg{catch22} and \code{fluctanal\_prop\_r1} from \pkg{tsfeatures}). 
However, interpreting this table is challenging as the relationships between the features are unknown---are all the 40 features behaving differently, or are they all highly correlated to each other and essentially proxy metrics for the same underlying time-series property?
We can better understand these relationships by visualizing the pairwise feature $\times$ feature correlation matrix.

\begin{table}[ht]
\centering
\begin{adjustbox}{max width=\textwidth}
\begin{tabular}{rllrr}
  \hline
 Feature & Feature Set & Classification Accuracy & $p$-value \\ 
  \hline
SC\_FluctAnal\_2\_rsrangefit\_50\_1\_logi\_prop\_r1 & catch22 & 53.8\% & $p$ $<$ .001 \\ 
  fluctanal\_prop\_r1 & tsfeatures & 53.8\% & $p$ $<$ .001 \\ 
  0\_wavelet\_energy\_7 & TSFEL & 53.8\% & $p$ $<$ .001 \\ 
  0\_wavelet\_standard\_deviation\_7 & TSFEL & 53.8\% & $p$ $<$ .001 \\ 
  0\_wavelet\_energy\_5 & TSFEL & 52.0\% & $p$ $<$ .001 \\ 
  0\_wavelet\_standard\_deviation\_5 & TSFEL & 52.0\% & $p$ $<$ .001 \\ 
  0\_wavelet\_energy\_4 & TSFEL & 51.4\% & $p$ $<$ .001 \\ 
  0\_wavelet\_standard\_deviation\_4 & TSFEL & 51.4\% & $p$ $<$ .001 \\ 
  0\_wavelet\_energy\_8 & TSFEL & 51.0\% & $p$ $<$ .001 \\ 
  0\_wavelet\_standard\_deviation\_8 & TSFEL & 51.0\% & $p$ $<$ .001 \\ 
  values\_autocorrelation\_lag\_6 & tsfresh & 50.6\% & $p$ $<$ .001 \\ 
  0\_wavelet\_energy\_6 & TSFEL & 50.6\% & $p$ $<$ .001 \\ 
  0\_wavelet\_standard\_deviation\_6 & TSFEL & 50.6\% & $p$ $<$ .001 \\ 
  values\_autocorrelation\_lag\_7 & tsfresh & 50.4\% & $p$ $<$ .001 \\ 
  seas\_acf1 & Kats & 50.4\% & $p$ $<$ .001 \\ 
  values\_agg\_linear\_trend\_attr\_stderr\_chunk\_len\_10\_f\_agg\_mean & tsfresh & 50.0\% & $p$ $<$ .001 \\ 
  hurst & Kats & 50.0\% & $p$ $<$ .001 \\ 
  0\_wavelet\_variance\_6 & TSFEL & 49.8\% & $p$ $<$ .001 \\ 
  0\_wavelet\_variance\_7 & TSFEL & 49.6\% & $p$ $<$ .001 \\ 
  0\_spectral\_distance & TSFEL & 49.4\% & $p$ $<$ .001 \\ 
  values\_autocorrelation\_lag\_5 & tsfresh & 48.8\% & $p$ $<$ .001 \\ 
  0\_root\_mean\_square & TSFEL & 48.8\% & $p$ $<$ .001 \\ 
  0\_area\_under\_the\_curve & TSFEL & 48.8\% & $p$ $<$ .001 \\ 
  0\_peak\_to\_peak\_distance & TSFEL & 48.6\% & $p$ $<$ .001 \\ 
  values\_agg\_linear\_trend\_attr\_stderr\_chunk\_len\_5\_f\_agg\_max & tsfresh & 48.4\% & $p$ $<$ .001 \\ 
  values\_maximum & tsfresh & 48.4\% & $p$ $<$ .001 \\ 
  values\_agg\_linear\_trend\_attr\_stderr\_chunk\_len\_5\_f\_agg\_mean & tsfresh & 48.4\% & $p$ $<$ .001 \\ 
  values\_agg\_linear\_trend\_attr\_stderr\_chunk\_len\_10\_f\_agg\_max & tsfresh & 48.4\% & $p$ $<$ .001 \\ 
  values\_agg\_linear\_trend\_attr\_stderr\_chunk\_len\_50\_f\_agg\_var & tsfresh & 48.4\% & $p$ $<$ .001 \\ 
  0\_wavelet\_variance\_8 & TSFEL & 48.4\% & $p$ $<$ .001 \\ 
  0\_max & TSFEL & 48.4\% & $p$ $<$ .001 \\ 
  0\_mean\_absolute\_deviation & TSFEL & 48.2\% & $p$ $<$ .001 \\ 
  shift\_level\_max & feasts & 48.0\% & $p$ $<$ .001 \\ 
  var\_tiled\_mean & feasts & 47.8\% & $p$ $<$ .001 \\ 
  stability & tsfeatures & 47.8\% & $p$ $<$ .001 \\ 
  values\_agg\_linear\_trend\_attr\_intercept\_chunk\_len\_50\_f\_agg\_min & tsfresh & 47.8\% & $p$ $<$ .001 \\ 
  values\_agg\_linear\_trend\_attr\_stderr\_chunk\_len\_10\_f\_agg\_min & tsfresh & 47.8\% & $p$ $<$ .001 \\ 
  values\_standard\_deviation & tsfresh & 47.6\% & $p$ $<$ .001 \\ 
  values\_linear\_trend\_attr\_stderr & tsfresh & 47.6\% & $p$ $<$ .001 \\ 
  0\_standard\_deviation & TSFEL & 47.6\% & $p$ $<$ .001 \\ 
   \hline
\end{tabular}
\end{adjustbox}
\caption{\label{tab:top_feature_table}
    \textbf{Comparison of classification accuracy and $p$-values between the top 40 individual features in \pkg{theft} for the Bonn EEG dataset}.
    $p$-values were calculated by comparing each individual feature's model classification accuracy against a Gaussian null distribution parameterized by the mean and \textit{SD} of model-free shuffled samples.
    These values were generated by the code in Listing~\ref{lst:sf_classifier}.
    }
\end{table}

\begin{figure}[t!]
  \centering
  \includegraphics{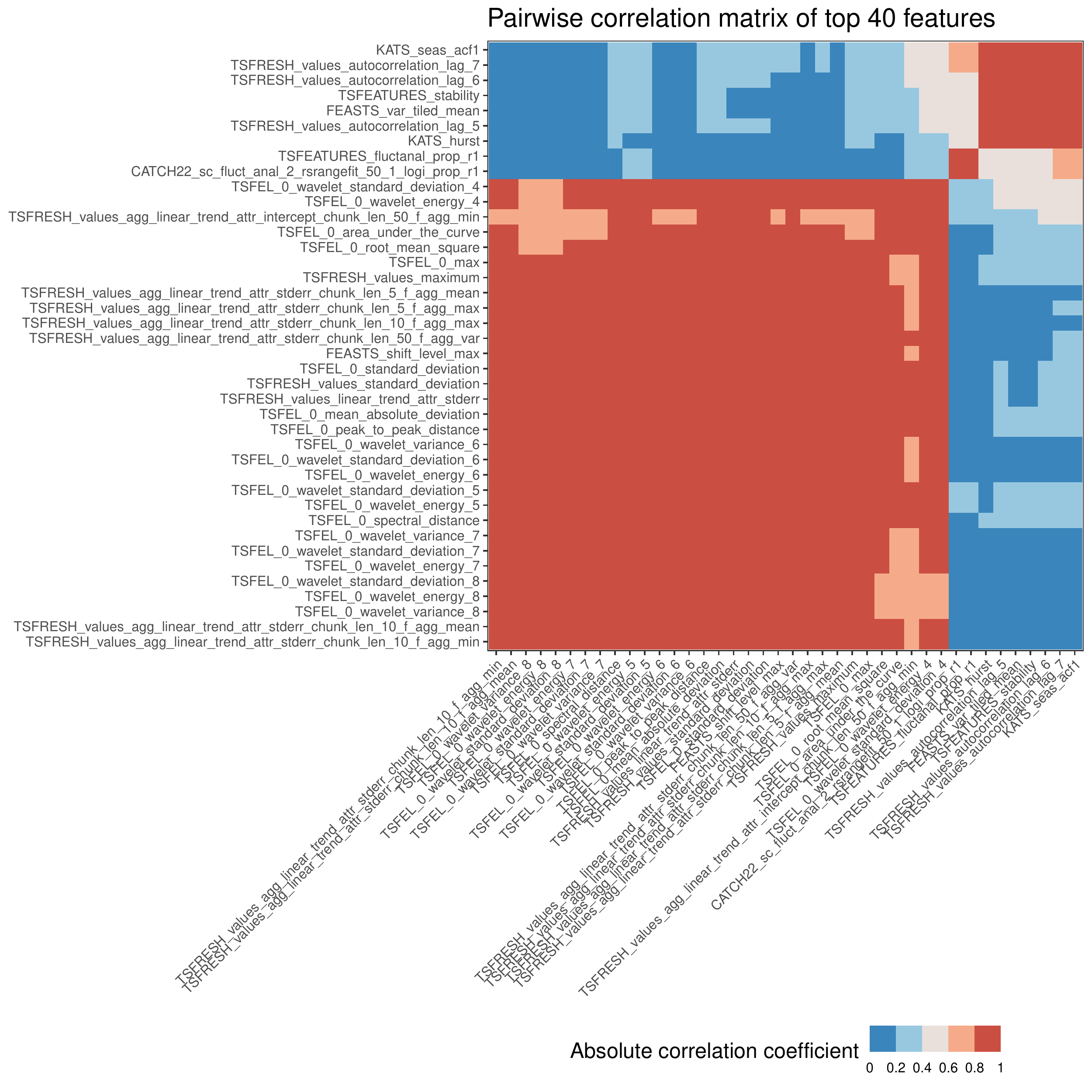}
  \caption{\label{fig:top-feature-correlations}
  \textbf{A group of variance-sensitive features and a group of autocorrelation-sensitive features perform the best at distinguishing between the five classes in the Bonn EEG dataset using the absolute Spearman correlation coefficient, $|\rho|$, to capture feature--feature similarity}.
  To aid the identification of similarly performing features, the matrix of correlation coefficients between features were then organized using hierarchical clustering (on Euclidean distances with average linkage) along rows and columns to order the heatmap graphic.
  This plot was generated by the code in Listing~\ref{lst:sf_classifier}.
  }
\end{figure}

The plot of the pairwise absolute correlation coefficients between the top 40 features (returned as \code{FeatureFeatureCorrelationPlot} in Listing~\ref{lst:sf_classifier}) is displayed in Fig.~\ref{fig:top-feature-correlations}.
The plot reveals two main groups of highly correlated ($|\rho| \gtrapprox 0.8$) features: in the bottom left and upper right of the plot.
The large cluster in the bottom left (containing features from \pkg{tsfresh}, \pkg{TSFEL}, and \pkg{feasts}) contains features sensitive to signal variance.
While a number of features in this cluster have names associated with wavelets (e.g., \code{TSFEL\_0\_wavelet\_variance\_7}), the plot reveals that on this dataset, these features exhibit similar behavior as features measuring signal variance (e.g., \code{TSFEL\_0\_standard\_deviation}).
The second cluster, in the top right, contains features that capture different types of autocorrelation structure in the time series, including linear autocorrelation coefficients (e.g., \code{TSFRESH\_values\_autocorrelation\_lag\_7} and \code{KATS\_seas\_acf1}).
The smaller third sub-cluster (a sub-group of the autocorrelation cluster) is comprised of two copies of the exact same feature as described earlier---\\\code{SC\_FluctAnal\_2\_rsrangefit\_50\_1\_logi\_prop\_r1} (which is contained in both \pkg{catch22} and \pkg{tsfeatures}).
This analysis reveals that the two broad types of time-series properties that best distinguish the classes of the Bonn EEG dataset are in fact very simple: variance and linear autocorrelation structure.
This understanding was obtained by comparing the results across six different open-source feature sets, aided by the ability to inspect the table of top performing features alongside the clustered feature--feature correlation plot.
However, while the differences between classes in this case were simple, other, more complex features may perform the strongest on other problems, or even different pairs of classes within the five-class dataset investigated here.
Identifying when simple features perform well is important as it can provide interpretable benchmarks for assessing relative performance gains achieved by more complex and/or less interpretable alternative classifiers.

\begin{figure}[t!]
  \centering
  \includegraphics{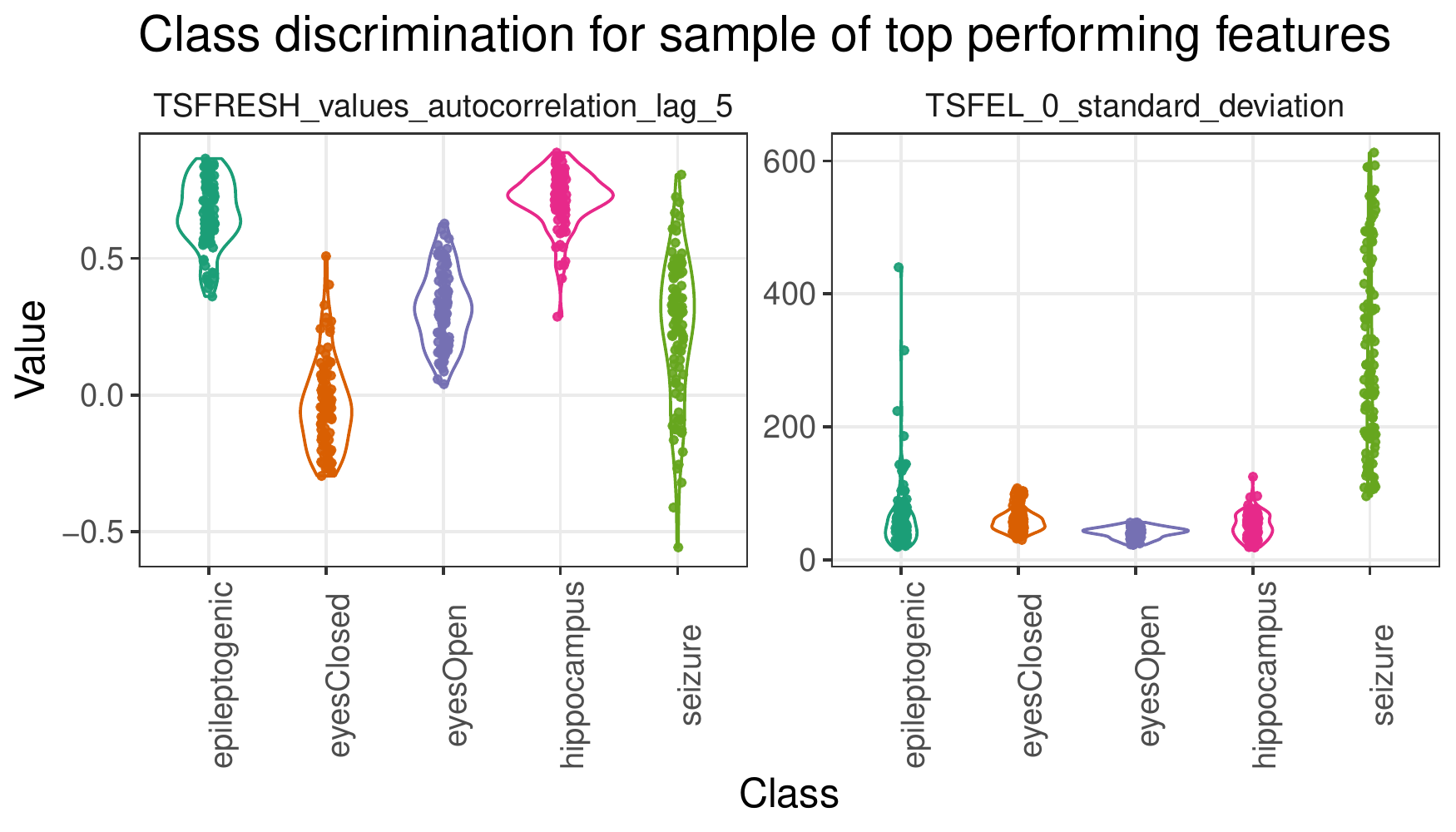}
  \caption{\label{fig:top-feature-violins}
  \textbf{Violin plots (on original feature value scale) of a sample of two of the top 40 features of all six feature sets in \pkg{theft} for classifying Bonn EEG groups from the \code{compute\_top\_features} function}.
  Classes differ in their variance and autocorrelation properties.
  The features selected for this plot were determined through the code in Listing~\ref{lst:sf_classifier}.
  }
\end{figure}

Having identified the discriminative features, it is important understand how they differ amongst the labeled classes of a dataset.
This can be achieved by visualizing the distribution of values for each class for each of the features. 
In \pkg{theft}, \code{compute\_top\_features} produces the object \code{ViolinPlots}, where each time series is represented as a point colored by its class label.
Example code is shown in Listing~\ref{lst:sf_classifier}, which produces violin plots for all 40 top features.
Here, for visual clarity, we show the violin plots for a selected feature from the variance-sensitive cluster of features from Fig.~\ref{fig:top-feature-correlations}: \code{0\_standard\_deviation} from \pkg{TSFEL} (measures the standard deviation); and a selected feature from the autocorrelation-sensitive cluster of features: \code{values\_autocorrelation\_lag\_5} from \pkg{tsfresh} (calculates the autocorrelation coefficient at a time lag of 5 samples).
The outputs are shown in Fig.~\ref{fig:top-feature-violins}.
Consistent with their high classification scores, both features are informative of class differences.
The plot shows that with regards to autocorrelation structure, we see that `eyesClosed' exhibits the lowest coefficient at lag 5, while `hippocampus' and `epileptogenic' exhibit the highest.
The plot also highlights that `seizure' time series have increased standard deviation, consistent with prior work \citep{fulcherHighlyComparativeTimeseries2013}.

\subsection{Additional functionality}
\label{sec:additionalfuncs}

In addition to the functionality demonstrated here, \pkg{theft} includes a collection of other functions not demonstrated here, including visualizations of pairwise correlation matrices (of both feature vectors and raw time-series values), processing of \pkg{hctsa}-formatted \proglang{Matlab} files, and a number of functions for investigating and cleaning feature data.
These have been omitted from this article for space, but readers are encouraged to explore them in the detailed vignette that is included in \pkg{theft}, and in the source code \citep{theft_pkg}.

\subsection{Accompanying interactive web application}
\label{sec:application}

\begin{figure}[t!]
  \centering
  \includegraphics{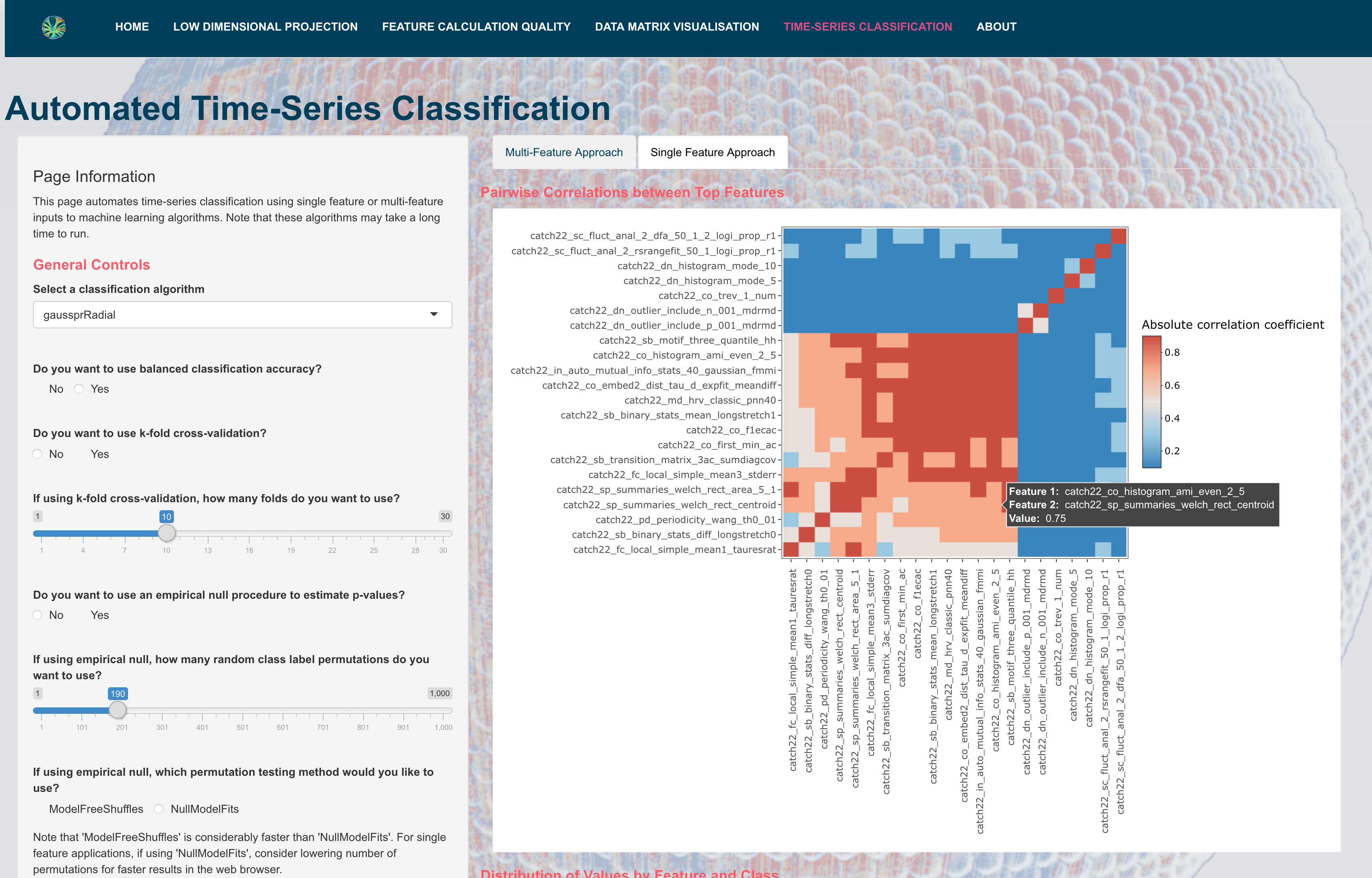}
  \caption{\label{fig:theft_webtool}
  \textbf{Example screenshot of the interactive web application implementation of \pkg{theft}}.
  An example of the interactivity of the top feature identification page is shown.
  The adjustable parameters on the left are user-interface renders of the function arguments in the \pkg{theft} package.
  }
\end{figure}

To provide the functionality of \pkg{theft} to analysts who may not be fluent with \proglang{R} and who also seek a fast impression of feature values and class discrimination, we have also developed an accompanying interactive web application \citep{theft_webtool} written in \pkg{Shiny} \citep{shiny}\footnote{\url{https://dynamicsandneuralsystems.shinyapps.io/timeseriesfeaturevis/}}.
The application allows users to upload a time-series dataset via a drag-and-drop interface, and the core functionality of \pkg{theft} can then be performed in the web browser.
Most of the graphical and computational functionality included in \pkg{theft} is presented in the web browser and users can download a file of the computed time-series features in a tidy format that can be read into any analysis program.
All the graphics available in \pkg{theft} are presented as interactive graphics by default to further enable an intuitive exploration of the uploaded dataset.
A screenshot of the informative feature identification page in the web application is displayed in Fig.~\ref{fig:theft_webtool}.

\section{Discussion}
\label{sec:discussion}

Feature-based time-series analysis is a powerful computational tool for solving problems using sequential (e.g., time-ordered) data.
We have introduced \pkg{theft}, an open-source package for \proglang{R} which implements the extraction, processing, visualization, and statistical analysis of time-series features.
The value of time-series features stems from their interpretability and strong connection to theory that can be used to understand empirical dynamics.
\pkg{theft} provides a unified interface to extracting features from six open-source packages---\pkg{catch22}, \pkg{feasts}, \pkg{tsfeatures}, \pkg{Kats}, \pkg{tsfresh}, and \pkg{TSFEL}---along with a comprehensive range of analyses to leverage the combined contributions from all of these packages.
For the first time in the free and open-source software setting, \pkg{theft} provides a full workflow for conducting feature-based time-series analysis, taking the analyst from feature extraction through to generating interpretable insights about their data.
We demonstrated \pkg{theft} on the five-class Bonn EEG time-series classification problem \citep{andrzejakIndicationsNonlinearDeterministic2001}, in which the full feature-based classification analysis pipeline---from feature extraction to normalization, classification, and interpretation of individual features---was achieved using a small number of key functions in \pkg{theft}.
\pkg{theft} can compare feature-set performance and leverage the combined set of features from all six packages, with in-built techniques like low-dimensional projections (\code{plot\_low\_dimension}) and feature--feature correlation matrices (\code{compute\_top\_features}) assisting in interpreting the patterns detected.
Analysts no longer need to construct complex workflows with multiple software libraries that were not designed to work together---\pkg{theft} provides a full suite of functionality, but also provides a blueprint for advanced users to alter and adapt as their research requires.

As new and more powerful features (and feature sets) are developed in the future, they can be incorporated into \pkg{theft} to enable ongoing assessments of the types of problems they are best placed to solve.
In addition to the analysis templates provided through functions in \pkg{theft}, there is much flexibility for users to adapt them or build new functionality for their own use-cases, such as applying different types of statistical learning algorithms on extracted feature matrices (e.g., feature selection), or to adapt the results to different applications such as extrinsic regression \citep{tanTimeSeriesExtrinsic2021} or forecasting \citep{montero-mansoFFORMAFeaturebasedForecast2020}.
Future work could also aim to reduce redundancy from across the combined features towards a new reduced feature set that combines the most generically informative and unique features from across the available feature-extraction packages (following the aims of the \pkg{catch22} feature set, selected from a library of $>7700$ candidate features in \pkg{hctsa} \cite{lubbaCatch22CAnonicalTimeseries2019}).

\section{Code Availability}
\label{sec:availability}

The source code for \pkg{theft} is available on GitHub at \url{https://github.com/hendersontrent/theft} \citep{theft_pkg}.

\bibliography{refs, mylibrary}

\end{document}